\documentclass[11pt,twocolumn]{article}

\usepackage{abstract}
\usepackage{amsmath}
\usepackage{authblk}
\usepackage[font=small]{caption}
\usepackage[hang,flushmargin]{footmisc}
\usepackage[a4paper,margin=0.98in]{geometry}
\usepackage{graphicx}
\usepackage{hyperref}
\usepackage{microtype}
\usepackage{natbib}
\usepackage{times}
\usepackage{titlefoot}
\usepackage{titlesec}
\usepackage{xcolor}
\usepackage{xurl}

\titleformat*{\section}{\large\bfseries}
\titleformat*{\subsection}{\normalsize\bfseries}
\titlespacing*{\section}{0pt}{12pt}{9pt}
\titlespacing*{\subsection}{0pt}{9pt}{6pt}

\hypersetup{
    colorlinks,
    linkcolor={blue!50!black},
    citecolor={blue!50!black},
    urlcolor={blue!80!black}
}

\setcitestyle{authoryear,open={(},close={)}}

\begin{document}

\title{\texorpdfstring{\vspace{-28pt}}{} \Large \bfseries CODEX: A Cluster-Based Method for Explainable Reinforcement Learning \texorpdfstring{\vspace{10pt}}{}}

\author[1]{\bfseries Timothy K. Mathes}
\author[2]{\bfseries Jessica Inman}
\author[1]{\bfseries Andr\'{e}s Col\'{o}n}
\author[3]{\bfseries Simon Khan}

\affil[1]{Assured Information Security, Inc.}
\affil[2]{Georgia Tech Research Institute}
\affil[3]{Air Force Research Laboratory}

\date{}

\maketitle

\begin{NoHyper}
    \unmarkedfntext{Presented at the International Joint Conference on Artificial Intelligence (IJCAI) 2023 Workshop on Explainable Artificial Intelligence (XAI).}
\end{NoHyper}

\begin{abstract}
\noindent Despite the impressive feats demonstrated by Reinforcement Learning (RL), these algorithms have seen little adoption in high-risk, real-world applications due to current difficulties in explaining RL agent actions and building user trust. We present Counterfactual Demonstrations for Explanation (CODEX), a method that incorporates semantic clustering, which can effectively summarize RL agent behavior in the state-action space. Experimentation on the MiniGrid and StarCraft II gaming environments reveals the semantic clusters retain temporal as well as entity information, which is reflected in the constructed summary of agent behavior. Furthermore, clustering the discrete+continuous game-state latent representations identifies the most crucial episodic events, demonstrating a relationship between the latent and semantic spaces. This work contributes to the growing body of work that strives to unlock the power of RL for widespread use by leveraging and extending techniques from Natural Language Processing.
\end{abstract}

\section{Introduction}

Reinforcement Learning (RL) is a revolutionary technology capable of superhuman long-term decision-making in complex and fast-paced domains \citep{tesauro92,mnih15,silver18,schrittwieser20,vinyals19}. Effective RL-enabled systems will readily outperform the greatest human minds at most tasks.\footnotemark However, a major challenge in the field has been explaining RL agent decisions. This prohibitive limitation is because existing Explainable Reinforcement Learning (XRL) methods do not effectively account for the fact that autonomous decision-making agents can change future observations of data based on actions they take or effectively reason over long-term objectives of the underlying agent mission. For example, AI AlphaStar competes against top-tier StarCraft II players, but gaining an understanding of the AI requires extensive empirical study. Effective XRL approaches that overcome these limitations are necessary to unlock the power of RL for widespread use.

\footnotetext{\url{https://www.nitrd.gov/pubs/National-AI-RD-Strategy-2019.pdf}}

\begin{figure}[t]
    \centering
    \includegraphics[width=\columnwidth]{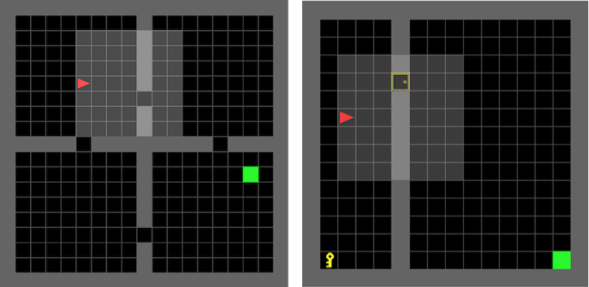}
    \caption{Four Rooms and Door Key MiniGrid environments. The RL agent (red triangle) is tasked with autonomously reaching the goal (green square) by maneuvering walls and locked doors.}
    \label{figure:1}
\end{figure}

One potential approach is to develop text-based XRL techniques using world models \citep{hafner20}. World models have proven to be extremely effective as several of the recent top-performing RL algorithms are world modeling based \citep{kaiser20,schrittwieser20,schrittwieser21,hafner21}. They may be used to show a user: a) what the RL agent expects is happening after it makes a decision; and b) what the RL agent expects would have happened had it made a different decision. The former is termed a \emph{factual} and the latter a \emph{counterfactual}.

In this paper, we propose a global post-hoc clustering method for XRL called Counterfactual Demonstrations for Explanation (CODEX) as a step towards understanding factuals and counterfactuals. CODEX automatically produces natural language episode tags that describe the agent states and actions while interacting with MiniGrid \citep{chevalier18} and StarCraft II \citep{vinyals17} environments. \autoref{figure:1} shows two MiniGrid environments in which the agent must navigate walls and locked doors to reach the goal.

CODEX offers several benefits: 1) the vector representations are densely clustered with very good separation, even when the tags are short ($\sim$5-6 words) with minimal semantic distinctiveness; 2) the centroid conditioned cluster topics are fully extractive, avoiding the issue of hallucinations observed in SOTA summarization models \citep{kryscinski19,falke19}; 3) the semantic clusters retain temporal as well as entity information, which in turn is reflected in the constructed summary of agent behavior in the state-action space; 4) two user-defined parameters provide more fine-grained and detailed summaries, revealing tags that occur rarely and may be important; and 5) clustering discrete+continuous game-state latent representations visually reveals the most crucial episode tags, demonstrating a relationship between the latent and semantic spaces.

By summarizing world model based factual and counterfactual examples while combining them with latent cluster visualizations, our method enables an intuitive and broader understanding of an RL agent's behavior. Our code is publicly available.\footnote{\url{https://github.com/ainfosec/codex}}

\section{Related Work}

\textbf{Explainability in Reinforcement Learning.} While Explainable Artificial Intelligence (XAI) has established and widely accepted techniques such as the SHAP library \citep{lundberg17} and its encompassed methods \citep{ribiero16,strumbelj14,shrikumar17,datta16,bach15,lipovetsky01}, XRL research has not yet yielded such well-regarded methods.

To aid in the development of new XRL approaches, XRL researchers have created useful taxonomies to describe and compare methods. There is a 2-step taxonomy largely established from \citep{puiutta20}, based upon XAI taxonomy \citep{adadi18}, that incorporates ideas from \citep{heuillet21}, based upon XAI taxonomy \citep{arrieta20}: scope and extraction type. Explanation scope can be either local or global while explanation extraction types are either intrinsic or post-hoc. Local explanations provide insight into specific predictions while global explanations provide insight into overall model structure, or logic.

Several recent works are related to, yet distinct from, CODEX. \cite{vanderwaa18} leverages a policy simulation and a translation of states and actions to a description that is easier to understand for human users to enable an RL agent to explain its behavior with contrastive explanations and in terms of the expected consequences of state transitions and outcomes. This work discerns interpretable state changes by applying classification models to the state representation. In contrast, CODEX leverages decoded visual representations of states to identify semantic properties.

Nguyen et al.\ leverage human annotations of interpretations of agent behavior along with automated rationale generation to create natural language explanations for a sequence of actions taken by an RL agent \citep{nguyen22}. This work is closely related to CODEX in that observed agent behaviors are summarized with text but differs in its requirement for a large set of human annotations. CODEX does not require human annotation of observations in order to generate a text summary of an episode. Additionally, Nguyen et al.\ describe a human's expectation of why an agent may take a particular action. This type of human bias works well in real-world scenarios where humans have a good understanding of world dynamics. CODEX, in contrast, leverages the RL agent's world model based understanding of its environment, which enables CODEX to elucidate deficiencies in the agent's understanding of its world.

\section{The MiniGrid and StarCraft II Environments} \label{section:environments}

MiniGrid refers to a collection of simple and easily configurable grid-world environments designed for RL research \citep{chevalier18}. The games feature a triangle-shaped player that must reach the goal in a discrete action space depending on the type of game. StarCraft II is a real-time strategy game that involves fast paced micro-actions as well as high-level planning and execution. For this effort, we use Deepmind's PySC2 API \citep{vinyals17} to train models to play two StarCraft II minigames, which provide simplified environments and objectives for the agents to learn.

We implement the world model based DreamerV2 \citep{hafner21} RL agent to enable visualization of counterfactuals. DreamerV2 uses a Recurrent State Space Model (RSSM) as its dynamics model to predict transitions. It accepts an encoded state, $z_t$, represented as a latent feature vector that encodes observable and inferred state information and an action, $a_t$, as input to predict what state the world will transition to, $z_{t+1}$, and the reward, $\hat{r}_{t+1}$, that will be received as the output. To reduce the complexity of the RSSM model, DreamerV2 uses a paired encoder-decoder Variational Auto-Encoding (VAE) architecture to learn how to encode raw high-dimensional observed states, $x$, as low-dimensional feature vectors, $z$. We term the latent representation $z$ a Dreamer state. The entire DreamerV2 architecture is trained in a self-supervised manner using sequences of observational data samples from the target domain, i.e., episodes, by comparing the predicted transitions, $\left[\hat{x}_{t+1}, \hat{r}_{t}\right]$, with actual observed transitions, $\left[x_{t+1}, r_t\right]$. The trained decoder can project an image of a predicted world-state from any plausible latent state, $z$, which is central to how CODEX visualizes counterfactuals.

\section{Framework} \label{section:framework}

CODEX generates natural language tags for MiniGrid and StarCraft II based primarily on the locations (i.e., coordinates) of the entities in the game. Before we can annotate MiniGrid episodes with natural language, we need to know which of the four directions (left, right, up, or down) the player faces at each timestep. Fortunately, the player's triangular shape means we can compute its direction using image moments, i.e., the weighted averages of the player's pixel intensities. For a greyscale image with pixel intensities $I(x, y)$, the raw image moments $M_{pq}$ are given by:
\begin{equation*}
    M_{pq} = \sum_x \sum_y x^p y^q I(x, y)
\end{equation*}
Likewise, the central moments $\mu_{pq}$ are given by:
\begin{equation*}
    \mu_{pq} = \sum_x \sum_y \left(x - \bar{x}\right)^p \left(y - \bar{y}\right)^q I(x, y)
\end{equation*}
We can discern whether the player is facing one of up/down or one of left/right using first- and second-order central moments:
\begin{equation*}
    \Theta = \frac{1}{2} \arctan \left(\frac{2\mu'_{11}}{\mu'_{20} - \mu'_{02}}\right)
\end{equation*}
where
\begin{align*}
    \mu'_{20} &= \mu_{20} / \mu_{00} = M_{20} / M_{00} - \bar{x}^2 \\
    \mu'_{02} &= \mu_{02} / \mu_{00} = M_{02} / M_{00} - \bar{y}^2 \\
    \mu'_{11} &= \mu_{11} / \mu_{00} = M_{11} / M_{00} - \bar{x}\bar{y}
\end{align*}
Once the entities' coordinates and player's direction are extracted, we generate tags using templates corresponding to specific states or events. For MiniGrid, we use:

\begin{center}
    \begin{tabular}{l}
        \hline
        The player/goal/key/door is at $(x,y)$. \\
        The player is facing left/right/up/down. \\
        The player turns left/right. \\
        The player moves forward. \\
        The player reached the goal. \\
        The door is open/closed. \\
        The key has been picked up. \\
        The player picks up/drops the key. \\
        The player opens/closes the door. \\
        \hline
    \end{tabular}
\end{center}

\noindent Because a DreamerV2 world model can sometimes produce invalid images from a state vector, we include two additional tags to annotate visual anomalies:

\begin{center}
    \begin{tabular}{l}
        \hline
        The player is facing a non-cardinal direction. \\
        The state of the door is unknown. \\
        \hline
    \end{tabular}
\end{center}

\noindent Note that the set of templates includes both state-driven tags, e.g., ``The key is at $(x, y)$.''\ which appear at every timestep, and event-driven tags like ``The player picks up the key.''\ which appear only at the timestep where a specific action is taken. For StarCraft II, we use the following:

\begin{center}
    \setlength{\tabcolsep}{4pt}
    \begin{tabular}{l}
        \hline
        Marine/Beacon [ID] moves from $(x{,}y)$ to $(x{,}y)$.\\
        Beacon/Shard appears at $(x{,}y)$. \\
        Shard [ID] is collected. \\
        Marine [ID] collects shard [ID]. \\
        Marine [ID] moves closer to/farther from \\
        group/shard [ID]. \\
        \hline
    \end{tabular}
\end{center}

\noindent where each entity has a unique, randomly generated [ID]. Additionally, we use the following group-related tags:

\begin{center}
    \begin{tabular}{l}
        \hline
        Entity [ID] leaves/joins group [ID]. \\
        Entities [IDs] leave/join/form group [ID]. \\
        Group [ID] is dissolved. \\
        Group [ID] merges with group [ID]. \\
        Group [ID] moves from $(x, y)$ to $(x, y)$. \\
        \hline
    \end{tabular}
\end{center}

We now present our summarization pipeline, constructed from 3 key components: 1) the contextualized embeddings language model; 2) the dimensionality reduction and clustering algorithms; and 3) the topic model.

We evaluate three pretrained Transformer-based language models: BERT-base \citep{devlin18}, BERTweet-base \citep{nguyen20}, and paraphrase-MiniLM-L6-v2 \citep{reimers19}. We considered BERTweet because the pre-training data includes short text and MiniLM because the fine-tuning on paraphrase detection may be advantageous on data with minimal semantic differences. We then employ the UMAP algorithm \citep{mcinnes18} for dimensionality reduction followed by HDBSCAN \citep{campello13} for semantic cluster enumeration.

The game-state summary $S'$ is constructed by identifying the most important tags in each semantic cluster. This is done by incorporating a Latent Dirichlet Allocation (LDA) \citep{blei03} topic model to predict the exemplar tag for each cluster. LDA's ngram\_range parameter is set to the length of the shortest and longest tags in a cluster. The underlying idea is that entire tags may be selected as the exemplar for each cluster, avoiding the issue of hallucinations by being fully extractive. However, there are cases when LDA predicts a substring to be the cluster topic because of how ngram\_range is set. For instance, ``Marine [ID] moves closer to beacon \_'' which is missing the beacon ``[ID].'' In these cases, the tag closest to the cluster centroid that contains the LDA topic substring is selected. In addition, cosine similarities between the selected tag vector and the rest of the tag vectors in the cluster are calculated. If the similarity falls below a threshold of 0.6, the corresponding tag is selected as well. We re-use the 0.6 parameter setting from previous semantic similarity work. Presumably, such tags increase the summary's informativeness. The selected tags are sorted by step number to produce the final summary $S'$.

\section{Evaluation Metrics} \label{section:metrics}

To the best of our knowledge, datasets with gold summaries for XRL on gaming environments are not publicly available, making automatic evaluation a challenge. In order to evaluate clustering performance on the language model embeddings, we adopt two metrics given the lack of ground truth labels.

\textbf{Silhouette Coefficient.} The calculation for a single sample $s$ consists of the mean distance between a sample and all other points in the same cluster $a$, and the mean distance between the sample $s$ and all other points in the next nearest cluster $b$. It is a measure of how well defined the clusters are spatially \citep{rousseeuw87}. The metric is defined as: $s = (b - a) \div \max(a, b)$.

\textbf{Global Cosine Similarity.} We take the mean across all clusters of each cluster's cosine similarity, which is computed as the mean cosine similarity between all cluster vectors and the cluster's centroid vector. It is a measure of semantic homogeneity and density:
\begin{equation*}
    \frac{1}{m} \sum_{i=1}^m \left(\frac{1}{n} \sum_{j=1}^n \frac{A_j \cdot v_c}{\left\lVert A_j \right\rVert \left\lVert v_c \right\rVert} \right)_i
\end{equation*}
where $m$ is the number of clusters, $n$ is the number of cluster vectors, $A_j$ is a given cluster vector, and $v_c$ is the cluster centroid vector.

\begin{table*}[ht]
    \centering
    \small
    \begin{tabular}{l c c c c c c}
        \hline
        \textbf{model} & \textbf{n\_neighbors} & \textbf{min\_cluster\_size} & \textbf{clustered (\%)} & \textbf{sil\_score} & \textbf{global\_cos\_sim} & \textbf{mean} \\
        \hline
        BERT-base & 10 & 10 & 98.8 & 0.916 & 0.986 & 0.951 \\
        BERTweet & 10 & 10 & 98.4 & 0.913 & 0.993 & \textbf{0.953} \\
        MiniLM & 10 & 10 & 98.6 & 0.902 & 0.964 & 0.933 \\
        \hline
    \end{tabular}
    \caption{BERT-base, BERTweet, and MiniLM peak performance on the MiniGrid-100 episodes.}
    \label{table:1}
\end{table*}

\section{Experiments}

To analyze the effectiveness of our CODEX method from different perspectives, we propose three research questions (RQs) to guide our experiments:

\textbf{RQ1:} Which language model is the most appropriate choice for CODEX considering the tradeoffs between clustering performance and model efficiency in terms of size and inference time?

\textbf{RQ2:} Are the constructed summaries concise and informative while still retaining temporal and/or entity information from the state-action space?

\textbf{RQ3:} Does clustering an environment's DreamerV2 game-state latent representations reveal a relationship between the latent and semantic spaces?

\subsection{Semantic Clustering Comparison (RQ1)}

\hspace{\parindent} \textbf{Experimental Setup.} We experiment on 100 MiniGrid episodes (MiniGrid-100) for tags embedding, dimensionality reduction and semantic clustering to compare the performance of BERT-base, BERTweet, and MiniLM embeddings on short text. We do not conduct additional training or fine-tuning. We log the average amount of time it takes for each model to produce episode embeddings using an NVIDIA Quadro P1000 GPU card with 4 GB of memory. The choice of limited hardware is to assess whether CODEX could be used on edge devices with restricted resources. We undertake a sweep of two key parameters: UMAP's n\_neighbors $=$ \{10, 15, 20, 25, 30\} and HDBSCAN's min\_cluster\_size $=$ \{5, 10, 15, 20, 25\}. The remaining parameters are kept fixed: UMAP's min\_dist$=$0.0, n\_components$=$2, metric$=$cosine, and n\_epochs$=$500; HDBSCAN's min\_samples$=$1 and cluster\_selection\_method$=$leaf. These values are based on previous work. We compute the percentage of tags clustered as well, since HDBSCAN can identify datapoints as noise. It would be ideal to have the tags maximally clustered.

\textbf{Spatial Separation and Semantic Homogeneity.} To evaluate the semantic clusters, we take the mean of the Silhouette Coefficient (sil\_score) and Global Cosine Similarity (global\_cos\_sim) for each pair of parameter values across 100 episodes (i.e., n\_neighbors$=$10, min\_cluster\_size$=$5 for 100 episodes; n\_neighbors$=$10, min\_cluster\_size$=$10 for 100 episodes, etc.). \autoref{table:1} reports the best performing parameter values for each model. Full results are in \autoref{appendix:a}.

All three models perform the best when n\_neighbors$=$10 and min\_cluster\_size$=$10 on MiniGrid-100. The average number of tags clustered across all 100 episodes is in the range of 98.4-98.8\%. BERTweet has the highest sil\_score and global\_cos\_sim mean at 0.953, although BERT-base and MiniLM exhibit comparable performance at 0.951 and 0.933, respectively.

\textbf{Efficiency.} We consider model size and speed in the selection process as well. \autoref{table:2} shows model sizes along with the average amount of time (secs.) each model takes to generate episode embeddings when run on a single GPU.

\begin{table}[ht]
    \centering
    \begin{tabular}{l c c c c}
        \hline
        \textbf{model} & \textbf{\# params} & \textbf{dim.} & \textbf{\# epi.} & \textbf{mean} \\
        \hline
        BERT-base & 109M & 768 & 100 & 3.19 \\
        BERTweet & 135M & 768 & 100 & 0.337 \\
        MiniLM & 23M & 384 & 100 & \textbf{0.059} \\
        \hline
    \end{tabular}
    \caption{Model sizes, embed dimensions, and mean generation times (secs.) on the MiniGrid-100 episodes.}
    \label{table:2}
\end{table}

At 23M parameters, MiniLM produces episode embeddings in 0.059 seconds on average, across 100 episodes that contain a total of 15,392 natural language tags. This is significantly faster than the much larger BERT-base and BERTweet models. Moreover, we observe that the MiniLM semantic clusters are dense and well-separated (see \autoref{figure:2}).

We choose MiniLM for further experimentation to answer RQ2 and RQ3 given the smaller size, speed, and comparable clustering performance, while taking into account the environmental impact, financial cost, and other pitfalls associated with Large Language Models (see \citealp{bender21}; \citealp{wei22}; and \citealp{thompson22} for discussions).

\subsection{Summary Analysis (RQ2)}

Since the final summary is crucial to understanding agent behavior, we conduct an exploratory qualitative analysis to study: \emph{conciseness} -- do the summaries possess a sufficient number of tags that are interpretable by humans without redundancy? \emph{informativeness} -- do the summaries include all relevant information from the game-state space?

\begin{figure}[t]
    \centering
    \includegraphics[width=\columnwidth]{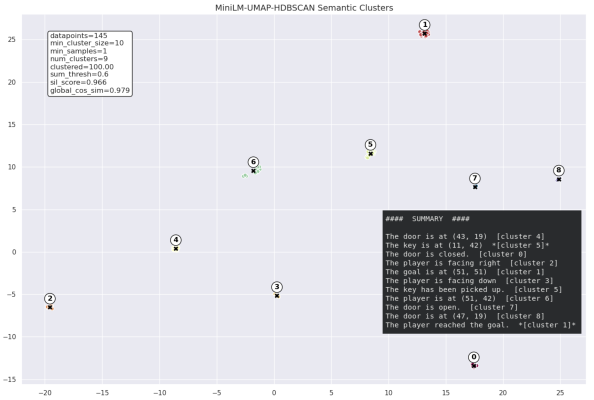}
    \caption{Semantic clusters and summary for MiniGrid episode 8.}
    \label{figure:2}
\end{figure}

\textbf{MiniGrid-100.} We visually inspect the semantic clusters and summaries from each MiniGrid-100 episode generated by the MiniLM $>>$ UMAP $>>$ HDBSCAN pipeline followed by LDA topic extraction. The summaries are constructed as outlined in \S\ref{section:framework}. \autoref{figure:2} is indicative of what we observe for the MiniGrid episodes. Labels are shown above each cluster with ``\textbf{x}'' marking the centroids.

As seen in the legend at the top left, episode 8 has 145 total event+state driven tags. The 9 enumerated clusters are dense and well-separated when min\_cluster\_size$=$10 and min\_samples$=$1 with 100\% of the tags clustered. The evaluation measurements approach a value of 1.0 with sil\_score$=$0.966 and global\_cos\_sim$=$0.979. The inset summary is shown to the right. Each line is formatted as: [tag] [cluster ID]. The asterisks around *[cluster ID]* indicate the tag is selected because it falls below the 0.6 summary threshold as explained in \S\ref{section:framework}. The summary consists of 11 total tags in this case, resulting in a low compression rate of $(11 \div 145) = 0.076$. Interestingly, clusters 0 and 7 are far apart in the UMAP 2-dim.\ projection even though the clustered tags differ by 1 word, i.e., ``The door is closed.''\ [cluster 0] vs.\ ``The door is open.''\ [cluster 7]. We hypothesize that MiniLM's fine-tuning on a paraphrase detection task enables it to generate separable embeddings when the semantic distinctions are minimal. The summary captures the key events during the game-state episode, such as ``The key has been picked up.''\ [cluster 5], ``The door is open.''\ [cluster 7], and ``The player reached the goal.''\ *[cluster 1]*.

\textbf{StarCraft II.} We visually inspect the semantic clusters and summaries from 100 randomly selected StarCraft II episodes out of 5,000 total episodes. These episodes have a significantly larger set of event-driven tags per episode compared to MiniGrid. Step numbers are included in the vector representations, designating when the events begin and end to consider the effect of providing the pipeline with temporal information. Each tag is prefixed with a timestamp denoting the starting and ending steps with ``$t_1$ {-}{-} $t_2$''. For instance, for the tag ``5 {-}{-} 8 Marine 4299161601 moves farther from shard 4299423745'', ``5'' is when the event starts (Step 5) and ``8'' is when it ends (Step 8).

\autoref{figure:4} in \autoref{appendix:b} illustrates what we typically observe in the StarCraft II episodes. We adjust min\_cluster\_size$=$20 and sum\_thresh$=$0.7, since the number of tags is approximately 3x larger. Episode 4 (\autoref{figure:4}) has 530 event-driven tags that are enumerated as 12 clusters when min\_cluster\_size$=$20. 93.58\% are clustered with tags marked as noise denoted in gray. The Silhouette Coefficient is 0.605, reflecting less cluster separation compared to MiniGrid. One possible reason is that the sequences of digits that make up the entity IDs place the vector representations close in 2-dim.\ space, although HDBSCAN predicts separate clusters.

\begin{figure}[t]
    \centering
    \includegraphics[width=\columnwidth]{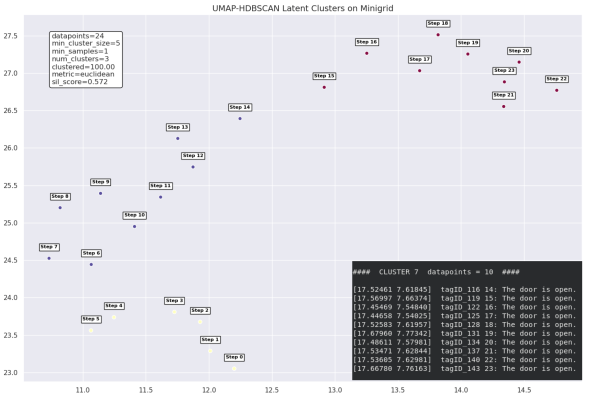}
    \caption{Latent representation clusters for MiniGrid episode 8.}
    \label{figure:3}
\end{figure}

It is notable that the clusters capture temporal as well as entity information. For example, ``13 {-}{-} 14 Marine 4298899457 moves closer to shard 4303880193'' [cluster 6] clusters separately from ``45 {-}{-} 48 Marine 4298899457 moves closer to shard 4303880193'' [cluster 9] when the only differences are the step numbers. In fact, we find that a cluster may have tags from the first half of the episode, which is reflected by the step numbers, with identical tags in another cluster from the second half. Consequently, the constructed summary includes exemplars from both halves of the episode. Moreover, we observe similar behavior when the differentiator is an entity ID. In terms of summary length, \autoref{figure:4} shows that the compression rate is low at $(15 \div 530) = 0.028$.

We visualize the clusters and summary from the same episode with min\_cluster\_size$=$10 (\autoref{figure:5} in \autoref{appendix:b}). The summary is longer and more fine-grained due to the enumeration of more clusters. [Cluster 0] at the bottom center is particularly interesting: it contains tags related to group behavior -- ``Entities 4299161601, 4298899457 form group 1'', ``Group 1 is dissolved'', ``Group 2 is dissolved'', ``Group 2 moves from (122, 128) to (110, 128)'', ``Group 3 is dissolved'', etc. While marked as noise when min\_cluster\_size$=$20 (\autoref{figure:4} at bottom center), the tags are clustered because the number of tags has met the threshold of min\_cluster\_size$=$10 in \autoref{figure:5}. We consider this significant because rare but important tags can be uncovered by adjusting min\_cluster\_size. Thus, users could adjust the parameters min\_cluster\_size and sum\_thresh along a sliding scale from course to fine-grained semantic clustering and summarization of an RL agent's behavior.

\subsection{Game-State Latent Representations (RQ3)}

We address RQ3 by clustering the DreamerV2 discrete+continuous latent representations from MiniGrid and StarCraft II with the UMAP $>>$ HDBSCAN piece of the pipeline. The 2048-dimensional, stepwise latent representations are extracted as described in \S\ref{section:environments} (i.e., the $z$ Dreamer states). \autoref{figure:3} shows latent clusters from MiniGrid-100 episode 8. These results are what we typically observe for MiniGrid episodes.

We find that lowering min\_cluster\_size$=$5 produces 3 clusters with 100\% of the 24 latent representations clustered. We see that Step 14 is the point in which ``The door is open.'' In the 2-dim.\ space, Step 14 operates as a transition point between its cluster and the next. In our careful inspection of the latent clusters from the MiniGrid episodes, we can surmise when the door opens by visually identifying the transition point. Moreover, we can predict when ``The key is picked up.''\ by identifying the last point between the first and second clusters (Step 5). The datapoints and their clusters form an arc from the bottom left to the top right for episode 8, which is a visual interpretation of the episode's progression through time.

In terms of StarCraft II, HDBSCAN did not cluster the latent representations, marking all points as noise for every episode. We hypothesize that MiniGrid's longer model training and lower-complexity environment are possible reasons why HDBSCAN successfully clusters the latent representations. An interesting future direction is exploring how much training is necessary to achieve latent representation clustering for a variety of environment complexities.

\section{Discussion}

In CODEX, we construct game-state summaries by identifying centroid-conditioned LDA topic exemplars for each semantic cluster. We choose this design over an abstractive summarization approach so that CODEX is an unsupervised method that is fully extractive in nature. We believe that this design choice contributes to user trust. Moreover, the finding that the MiniGrid latent representations can be clustered to reveal important episodic events prompts the scientific community to consider new research questions about the nature of the RL latent space. Ongoing research is investigating how to present factual and counterfactual summaries, show latent cluster visualizations, and allow for intuitive manipulation of the min\_cluster\_size and sum\_thresh parameters by users, which opens the door to understanding what an RL agent expects will happen vs.\ what would have happened had a different decision been made.

\section{Conclusion}

CODEX produces text-based summaries that provide representations of factuals and counterfactuals. These summaries could be leveraged to summarize collections of counterfactuals, perhaps with hierarchical summarization techniques. Other venues for future work include extending CODEX to additional semantically diverse environments, exploring the limits of CODEX with respect to episode length and state complexity, extracting more information from the latent space or increasing CODEX's efficiency by, e.g., automating state tagging with the latest computer vision techniques.

\onecolumn
\appendix

\section{Full Results on the MiniGrid-100 Episodes} \label{appendix:a}

\begin{table}[ht]
    \centering
    \begin{tabular}{c c c c c c}
        \hline
        \textbf{n\_neighbors} & \textbf{min\_cluster} & \textbf{clustered (\%)} & \textbf{sil\_score} & \textbf{global\_cos\_sim} & \textbf{mean} \\
        \hline
        10 & 5 & 93.8 & 0.747 & 0.992 & 0.870 \\
        \textbf{10} & \textbf{10} & \textbf{98.8} & \textbf{0.916} & \textbf{0.986} & \textbf{0.951} \\
        10 & 15 & 91.7 & 0.844 & 0.968 & 0.906 \\
        10 & 20 & 88.6 & 0.737 & 0.939 & 0.838 \\
        10 & 25 & 84.0 & 0.638 & 0.904 & 0.771 \\
        15 & 5 & 92.1 & 0.688 & 0.991 & 0.839 \\
        15 & 10 & 98.8 & 0.883 & 0.982 & 0.932 \\
        15 & 15 & 99.0 & 0.912 & 0.975 & 0.943 \\
        15 & 20 & 94.0 & 0.834 & 0.950 & 0.892 \\
        15 & 25 & 90.6 & 0.724 & 0.914 & 0.819 \\
        20 & 5 & 89.7 & 0.654 & 0.989 & 0.822 \\
        20 & 10 & 98.2 & 0.844 & 0.981 & 0.913 \\
        20 & 15 & 99.4 & 0.887 & 0.972 & 0.930 \\
        20 & 20 & 98.6 & 0.900 & 0.960 & 0.930 \\
        20 & 25 & 92.8 & 0.818 & 0.937 & 0.877 \\
        25 & 5 & 89.5 & 0.623 & 0.989 & 0.806 \\
        25 & 10 & 97.6 & 0.802 & 0.977 & 0.889 \\
        25 & 15 & 99.0 & 0.875 & 0.969 & 0.922 \\
        25 & 20 & 99.2 & 0.896 & 0.955 & 0.925 \\
        25 & 25 & 97.4 & 0.871 & 0.941 & 0.906 \\
        30 & 5 & 87.3 & 0.620 & 0.988 & 0.804 \\
        30 & 10 & 97.1 & 0.796 & 0.976 & 0.886 \\
        30 & 15 & 98.6 & 0.881 & 0.967 & 0.924 \\
        30 & 20 & 98.9 & 0.904 & 0.954 & 0.929 \\
        30 & 25 & 96.3 & 0.884 & 0.939 & 0.912 \\
        \hline
    \end{tabular}
    \caption{BERT-base performance on the MiniGrid-100 episodes.}
    \label{table:3}
\end{table}

\vfill
\newpage

\begin{table}[ht]
    \centering
    \begin{tabular}{c c c c c c}
        \hline
        \textbf{n\_neighbors} & \textbf{min\_cluster} & \textbf{clustered (\%)} & \textbf{sil\_score} & \textbf{global\_cos\_sim} & \textbf{mean} \\
        \hline
        10 & 5 & 93.8 & 0.764 & 0.996 & 0.880 \\
        \textbf{10} & \textbf{10} & \textbf{98.4} & \textbf{0.913} & \textbf{0.993} & \textbf{0.953} \\
        10 & 15 & 92.7 & 0.830 & 0.983 & 0.906 \\
        10 & 20 & 89.1 & 0.724 & 0.968 & 0.846 \\
        10 & 25 & 87.2 & 0.620 & 0.950 & 0.785 \\
        15 & 5 & 93.1 & 0.720 & 0.996 & 0.858 \\
        15 & 10 & 98.3 & 0.902 & 0.993 & 0.947 \\
        15 & 15 & 97.8 & 0.882 & 0.987 & 0.934 \\
        15 & 20 & 91.3 & 0.779 & 0.972 & 0.876 \\
        15 & 25 & 89.5 & 0.664 & 0.953 & 0.809 \\
        20 & 5 & 90.7 & 0.705 & 0.995 & 0.850 \\
        20 & 10 & 98.5 & 0.911 & 0.992 & 0.951 \\
        20 & 15 & 96.9 & 0.901 & 0.987 & 0.944 \\
        20 & 20 & 94.3 & 0.828 & 0.974 & 0.901 \\
        20 & 25 & 89.9 & 0.718 & 0.954 & 0.836 \\
        25 & 5 & 89.3 & 0.691 & 0.995 & 0.843 \\
        25 & 10 & 98.6 & 0.909 & 0.991 & 0.950 \\
        25 & 15 & 97.4 & 0.904 & 0.986 & 0.945 \\
        25 & 20 & 93.9 & 0.845 & 0.975 & 0.910 \\
        25 & 25 & 90.4 & 0.766 & 0.962 & 0.864 \\
        30 & 5 & 88.9 & 0.687 & 0.995 & 0.841 \\
        30 & 10 & 99.1 & 0.892 & 0.991 & 0.942 \\
        30 & 15 & 98.6 & 0.900 & 0.987 & 0.943 \\
        30 & 20 & 97.1 & 0.830 & 0.975 & 0.902 \\
        30 & 25 & 92.9 & 0.754 & 0.959 & 0.857 \\
        \hline
    \end{tabular}
    \caption{BERTweet performance on the MiniGrid-100 episodes.}
    \label{table:4}
\end{table}

\vfill
\newpage

\begin{table}[ht]
    \centering
    \begin{tabular}{c c c c c c}
        \hline
        \textbf{n\_neighbors} & \textbf{min\_cluster} & \textbf{clustered (\%)} & \textbf{sil\_score} & \textbf{global\_cos\_sim} & \textbf{mean} \\
        \hline
        10 & 5 & 93.1 & 0.719 & 0.983 & 0.851 \\
        \textbf{10} & \textbf{10} & \textbf{98.6} & \textbf{0.902} & \textbf{0.964} & \textbf{0.933} \\
        10 & 15 & 93.8 & 0.863 & 0.928 & 0.895 \\
        10 & 20 & 89.4 & 0.793 & 0.887 & 0.840 \\
        10 & 25 & 84.6 & 0.696 & 0.832 & 0.764 \\
        15 & 5 & 91.8 & 0.670 & 0.976 & 0.823 \\
        15 & 10 & 99.1 & 0.871 & 0.957 & 0.914 \\
        15 & 15 & 98.7 & 0.905 & 0.940 & 0.923 \\
        15 & 20 & 94.5 & 0.842 & 0.905 & 0.874 \\
        15 & 25 & 89.2 & 0.742 & 0.845 & 0.794 \\
        20 & 5 & 89.8 & 0.640 & 0.971 & 0.805 \\
        20 & 10 & 99.2 & 0.870 & 0.953 & 0.911 \\
        20 & 15 & 98.4 & 0.901 & 0.938 & 0.919 \\
        20 & 20 & 96.1 & 0.881 & 0.910 & 0.895 \\
        20 & 25 & 92.5 & 0.783 & 0.856 & 0.820 \\
        25 & 5 & 88.2 & 0.624 & 0.970 & 0.797 \\
        25 & 10 & 98.7 & 0.844 & 0.951 & 0.898 \\
        25 & 15 & 98.4 & 0.864 & 0.934 & 0.899 \\
        25 & 20 & 95.7 & 0.850 & 0.905 & 0.877 \\
        25 & 25 & 94.9 & 0.837 & 0.866 & 0.852 \\
        30 & 5 & 87.5 & 0.620 & 0.968 & 0.794 \\
        30 & 10 & 98.5 & 0.825 & 0.948 & 0.887 \\
        30 & 15 & 97.6 & 0.860 & 0.931 & 0.895 \\
        30 & 20 & 95.8 & 0.851 & 0.897 & 0.874 \\
        30 & 25 & 94.4 & 0.833 & 0.866 & 0.850 \\
        \hline
    \end{tabular}
    \caption{MiniLM performance on the MiniGrid-100 episodes.}
    \label{table:5}
\end{table}

\vfill
\newpage

\section{StarCraft II Example Semantic Clusters and Summaries} \label{appendix:b}

\begin{figure}[ht]
    \centering
    \includegraphics[width=6in]{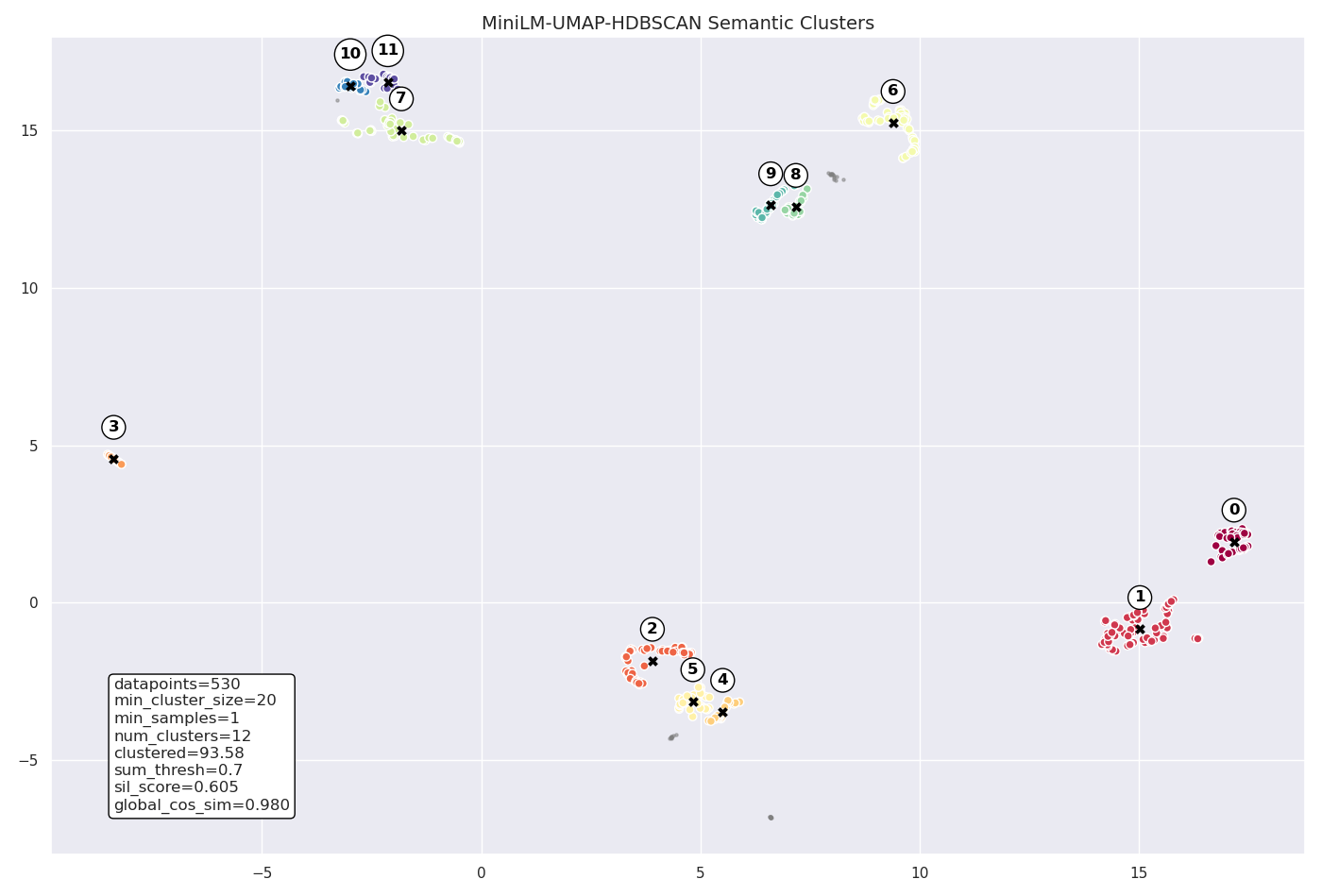}

    \vspace{1em}

    \includegraphics[width=4.32in]{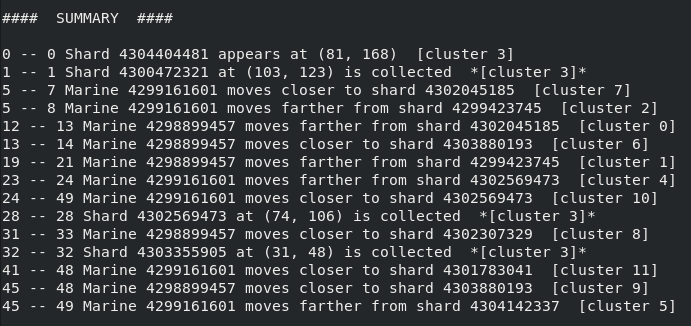}
    \caption{Semantic clusters and summary for StarCraft II episode 4 (min\_cluster\_size$=$20).}
    \label{figure:4}
\end{figure}

\vfill
\newpage

\begin{figure}[ht]
    \centering
    \includegraphics[width=6in]{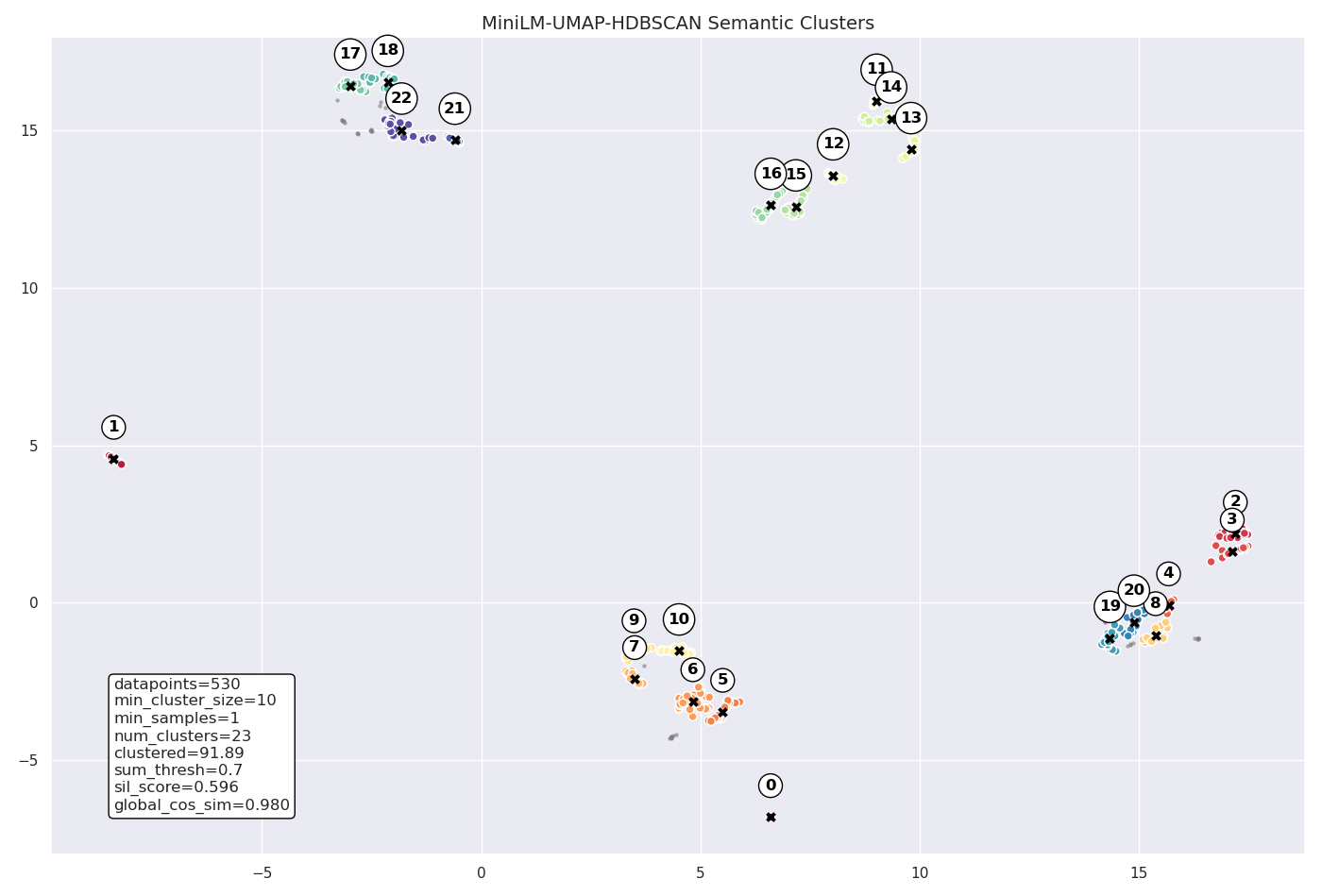}

    \vspace{1em}

    \includegraphics[width=4.39in]{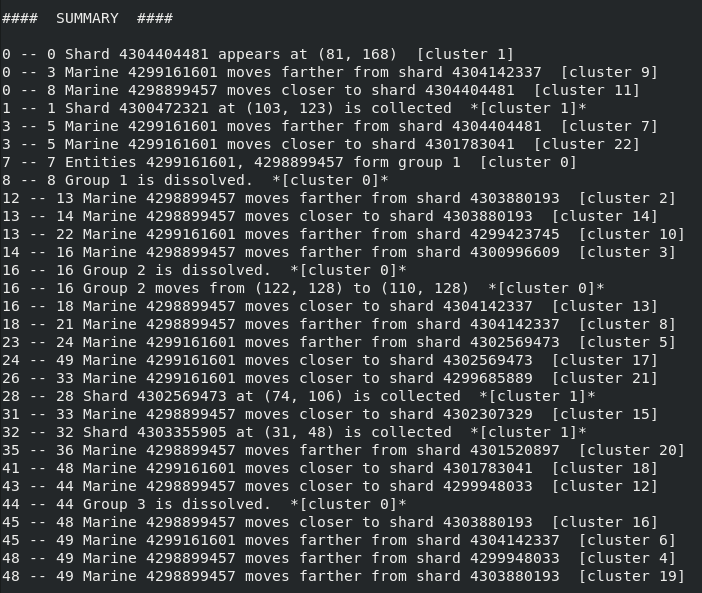}
    \caption{Semantic clusters and summary for StarCraft II episode 4 (min\_cluster\_size$=$10).}
    \label{figure:5}
\end{figure}

\vfill

\twocolumn
\section*{Ethical Statement}

Explainable Reinforcement Learning, as well as explainability more broadly, raises ethical concerns when applied to real-world problems. The main issue is the explanations could be used by malicious actors to manipulate AI systems, which would be especially pernicious for a field such as robotics.

For instance, consider a scenario where a robotic agent is sent to retrieve an apple as quickly as possible. There are two paths to the apple, one that is shorter and uses a set of stairs and one that is longer and uses an elevator. Suppose the agent chooses to retrieve the apple using the longer elevator path. A typical user may find this to be an unexpected choice since the longer path should take more time to traverse. By reviewing the counterfactual example, one can observe what the agent expects to happen if it chooses the shorter stair path. In this case, the agent expects to start a slow descent before falling down the stairs, finding itself unable to right itself and continue.

Given the above scenario, a malicious actor with access to the factual and counterfactuals would know that at least one of the counterfactuals causes a negative outcome. Thus, directing the robotic agent to choose the shorter path would result in its failure. This outcome would have far reaching consequences if the robotic agent is employed in a healthcare setting or in an area where dangerous materials are being handled. It is crucial to continue discussions on ways to mitigate this situation.

\section*{Acknowledgements}

This material is based upon work supported by the United States Air Force under Contract No.\ FA8750-22-C-1003. Any opinions, findings, and conclusions or recommendations expressed in this material are those of the author(s) and do not necessarily reflect the views of the United States Air Force.

\end{document}